%% file: neurips26-sde-mtsad.tex
\documentclass{article}

\usepackage[preprint]{neurips_2026} 
\usepackage[utf8]{inputenc}
\usepackage[T1]{fontenc}
\usepackage{booktabs}
\usepackage{amsfonts}
\usepackage{microtype}
\usepackage{xcolor}
\usepackage{graphicx}
\usepackage[acronym]{glossaries}
\usepackage{bibunits}
\usepackage{siunitx}

\usepackage{tabularx}
\usepackage[font=small,labelfont=bf]{caption}
\usepackage{subcaption}
\usepackage{float}
\usepackage{wrapfig}
\usepackage{placeins}
\usepackage{fontawesome5}
\usepackage{tikz}

\usepackage{hyperref}       
\hypersetup{
    colorlinks=true,
    linkcolor=blue,
    citecolor=blue,
    urlcolor=blue
}

\usepackage{cleveref}

\usepackage[table,dvipsnames]{xcolor}
\colorlet{first}{ForestGreen!50!white}
\colorlet{third}{ProcessBlue!12!white}
\colorlet{second}{first!35!third}

\usepackage{tikz}
\newcommand{\rankbox}[2][gray!10]{%
\tikz[baseline=(char.base)]{
  \node[
    rounded corners=2pt,
    draw=black,
    fill=#1,
    inner sep=1pt,
    font=\scriptsize,
    minimum width=0.8cm,
    minimum height=0.25cm,
    text width=0.4cm,
    align=right
  ] (char) {#2};
}}

\usepackage{xspace}
\newcommand{\std}[1]{\textcolor{gray}{\scriptsize\textpm\ #1}}
\newcommand{\Rn}{$\mathbb{R}^n$\xspace}
\newcommand{\Sn}{$\mathbb{S}^n$\xspace}

\defaultbibliographystyle{plainnat}
\newacronym{mtsad}{MTSAD}{Multivariate Time Series Anomaly Detection}
\newacronym{pca}{PCA}{Principal Component Analysis}
\newacronym{knn}{kNN}{k-Nearest Neighbors}
\newacronym{ocsvm}{OCSVM}{One-Class Support Vector Machine}
\newacronym{lof}{LOF}{Local Outlier Factor}
\newacronym{copod}{COPOD}{Copula-Based Outlier Detection}
\newacronym{iforest}{iForest}{Isolation Forest}
\newacronym{sde}{SDE}{Stochastic Differential Equation}
\newacronym{lsd}{LSD}{Latent SDE Anomaly Detection}

\newacronym{mtan}{mTAND}{multi-time attention network}
\newacronym{elbo}{ELBO}{evidence lower bound}
\newacronym{kl}{KL}{Kullback-Leibler}

\newacronym{msl}{MSL}{Mars Science Laboratory}
\newacronym{psm}{PSM}{Pooled Server Metrics}
\newacronym{qad}{QAPPD}{Quanser Aero 2 Pick-and-Place Dataset}
\newacronym{smap}{SMAP}{Soil Moisture Active and Passive}
\newacronym{smd}{SMD}{Server Machine Data}
\newacronym{swat}{SWaT}{Secure Water Treatment}
\newacronym{wadi}{WaDi}{Water Distribution}

\newcommand{\eg}{e.g.\xspace}
\newcommand{\ie}{i.e.\xspace}

\newcommand{\sthu}[1]{\marginpar{\tiny\textcolor{blue}{StHu: #1}}}
\newcommand{\fg}[1]{\marginpar{\tiny\textcolor{olive}{FG: #1}}}
\newcommand{\sthuinline}[1]{\textcolor{blue}{StHu: #1}}
\newcommand{\uram}[1]{\marginpar{\tiny\textcolor{red}{Martin: #1}}}
\newcommand{\todo}[1]{\textcolor{red}{\textbf{TODO: #1}}}
\newcommand{\ignore}[1]{}

\renewcommand{\sthu}[1]{}
\renewcommand{\sthuinline}[1]{}
\renewcommand{\fg}[1]{}
\renewcommand{\uram}[1]{}
\renewcommand{\todo}[1]{}

\title{Anomaly Detection for Sparse and Irregular Multivariate Time Series with Latent SDEs}

\author{%
    Martin Uray$^{1,2}$ \ \href{mailto:martin.uray@fh-salzburg.ac.at}{\textcolor{black}{\faEnvelope[regular]}}\, ,
    Florian Graf$^{2}$, Dominik Geng$^{2}$, Stefan Huber$^{1}$, Roland Kwitt$^{2}$ \\
    $^{1}$Josef Ressel Centre for Intelligent and Secure Industrial Automation,\\
    University of Applied Sciences, Salzburg, Austria \\
    $^{2}$University of Salzburg, Austria \\
}

\begin{document}

    \maketitle

    \begin{abstract}

        Multivariate time series anomaly detection (MTSAD) is critical for
        a wide range of application areas, such as industrial monitoring,
        cybersecurity, or healthcare. Real-world data is often sparse,
        irregularly sampled or partially observed, yet existing methods assume
        uniformly sampled time series.
        We propose a generative approach based on \emph{Latent SDEs} that
        projects the observed time series on a continuous-time stochastic
        dynamical system, directly being able to handle missing observations
        and irregular sampling, while also naturally capturing possible cyclic
        behavior that many real-world use cases inherently possess.
        Experiments on six anomaly benchmark datasets show that our proposed
        method ranks first among state-of-the-art baselines. We further
        demonstrate that our method remains robust under severe data sparsity,
        while performance significantly degrades for the tested baseline methods.
        These results highlight latent SDEs as a natural inductive bias for
        anomaly detection in multivariate time series, especially in presence
        of real-world irregularities.

    \end{abstract}

    \section{Introduction}
    \label{sec:introduction}

\begin{figure}[t]
            \centering
            \includegraphics[width=\linewidth]{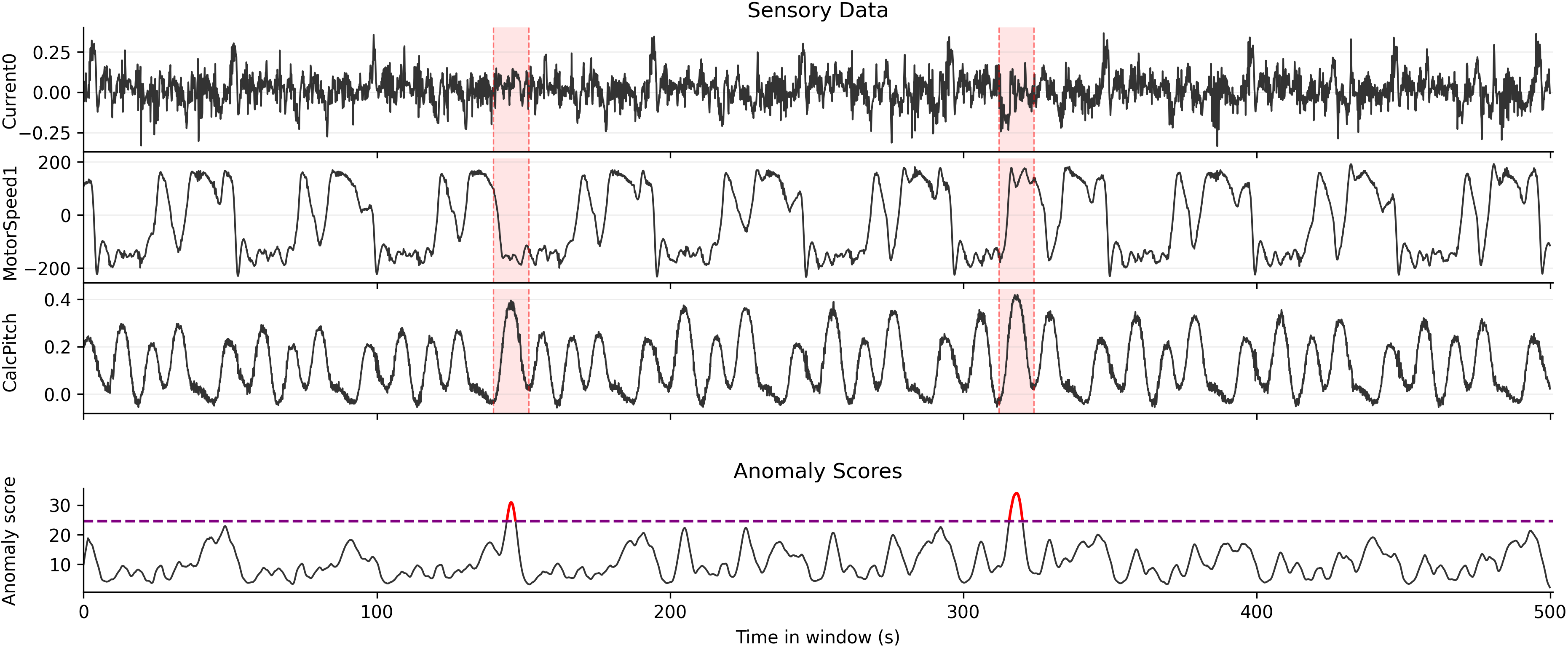}
            \caption{Anomaly detection on the \gls{qad} benchmark. \textbf{Top:} Three 
            representative features from trace~1 over a 500-second window, with 
            ground-truth anomalous regions shaded in red. \textbf{Bottom:} Corresponding 
            anomaly score computed as the log-likelihood under \gls{lsd}; detected 
            anomalies are shaded in red.}
            \label{fig:motivation-mtsad}
        \end{figure}

        Multi-variate time series data is ubiquitous in a variety of real-world
        applications, including industrial monitoring, cybersecurity, finance,
        or healthcare. Monitoring such data for anomalous patterns is critical
        for predictive maintenance in industry, fraud detection in finance,
        intrusion detection in cybersecurity, or early warning systems in
        healthcare and has been actively researched since the 1960s. With the
        increasing availability of data, the task of \gls{mtsad} gained
        attention in machine learning 
        research~\citep{blazquez2021,li2023,zamanzadeh2024}.
        Concerning the learning methodology, all paradigms from unsupervised to
        supervised, including semi- and self-supervised methods can be found in
        the literature. However, for many real-world applications, we lack labeled data, putting a focus on unsupervised methods. 
        Following this general scheme, we train a model exclusively on benign data; 
        anomalous patterns are then identified as out-of-distribution samples under 
        the learned model by thresholding an anomaly score.
        \Cref{fig:motivation-mtsad} illustrates this paradigm on three representative 
        features from the \gls{qad} benchmark, where the thresholded score yields a 
        binary prediction classifying each sample as either benign or anomalous.

        Furthermore, in many real-world applications, time series data is often
        noisy~\citep{cook2020}. More critical, however, it is often also highly
        irregular: Different variables are sampled unevenly, asynchronously or
        at different frequencies. Data might be sparse or partially missing.
        The prevalent industrial protocol OPC~UA implements data sparsity by
        design; data is only published when values deviate sufficiently from
        the last recorded value, cf.~\citep{opcua_part1}.
        %
        The assumption of uniformly sampled time series is therefore often
        unrealistic in practice, cf.~\citep{silva2012,weerakody2021}, yet
        state-of-the-art methods for \gls{mtsad} typically rely on this
        assumption.

        To still apply these methods, the data needs to be preprocessed, \eg,
        by imputation and resampling. However, this introduces biases and discards
        structural information, as we also confirm in our experiments.
        In line with \cite{belay2023}, who argue that such preprocessing should be 
        avoided altogether, we contend that \gls{mtsad} methods should natively 
        accommodate these data characteristics without relying on imputation or 
        resampling.

        On the other hand, such time series are often generated by some
        underlying dynamical process, whether it be a physical process in an
        industrial control system or a physiological process in healthcare.
        A particularly strong example is the machine and factory automation
        domain, \eg, pick-and-place tasks in robotics or injection molding
        machines, where the repetitive behavior of these machines leads to some
        form of roughly cyclic measurements in a geometric-topological sense
        \citep{SRMH25}. However, current state-of-the-art methods for
        \gls{mtsad} do not leverage this intrinsic structure.

        \ignore{Frame here the story line to a bit historical context: mtsad still
        young, data availability problem, jump to real-world problems and data.
        Now a chance for new problems.}
        From a historical perspective, only a handful of genuinely challenging 
        benchmarks exist for \gls{mtsad}: a large portion of available datasets have 
        been shown to be trivial to solve~\citep{wu2023a}, in the sense that 
        near-perfect detection can be achieved with a single line of code.
        More broadly, the 
        availability of a diverse set of real-world benchmark datasets remains limited, since industrial 
        data is frequently subject to privacy constraints and proprietary restrictions 
        that prevent public release~\citep{souza2020}. As a result, available benchmarks 
        originate primarily from three sources: internet companies that have identified 
        value in addressing their own operational anomaly detection 
        problems~\citep{su2019,abdulaal2021}, public organizations operating in 
        narrow, domain-specific fields of application~\citep{hundman2018}, and research 
        institutes collecting data from controlled testbed 
        scenarios~\citep{goh2017,chuadhry2017,nosrati2026}.
        These origins partly explain why the data challenges described above, namely 
        irregular sampling and natural sparsity, are largely absent from existing 
        benchmarks. 
        This work takes a step toward closing this gap by explicitly addressing these 
        real-world challenges in both the proposed method and its evaluation.

        \textbf{Contributions.} Against this background, we make the following 
        contributions:
        \begin{itemize}
            \item \textbf{A novel \gls{mtsad} method}, \gls{lsd}, that takes learning the 
                latent stochastic dynamics as a first principle. 
                This approach inherently accommodates irregularly
                sampled and sparse multivariate time series, without requiring
                imputation or resampling as a preprocessing step. We consider
                two latent manifolds, one of which is the hypersphere \Sn,
                which naturally accommodates cyclic behavior of the underlying
                process. 

            \item \textbf{Comprehensive benchmarking} across a diverse set of established 
                benchmarks, demonstrating consistently competitive performance against 
                classical and state-of-the-art baselines across all evaluated benchmark datasets.

            \item \textbf{Robustness under extreme sparsity.} Through a dedicated
                sparsity experiment reflecting data availability conditions commonly
                encountered in industrial deployments, we demonstrate that \gls{lsd}
                maintains stable detection performance, where state-of-the-art
                methods suffer substantial degradation, exhibiting drops of
                more than an order of magnitude smaller than those of the
                best-performing baseline.

        \end{itemize}

    \section{Related work}
    \label{sec:related-work}
    
    \textbf{Multivariate Time Series Anomaly Detection.}
    \gls{mtsad}~\citep{blazquez2021,li2023,zamanzadeh2024} is a complex and challenging problem, with substantial progress and numerous methods proposed in recent years.
    Existing work broadly falls into two categories: forecasting-based~\citep{xu2022,wu2023} and
    reconstruction-based~\citep{ruff2018,audibert2020,tuli2022} approaches.
    Forecasting-based methods learn the temporal dynamics of a time series and use 
    predictions over a given context window to anticipate future observations. The 
    deviation between the predicted and true measured values serves as an
    anomaly score. However, such approaches often struggle to generalise across
    multiple variates, as they typically model each variate independently,
    inherently overlooking spatial dependencies between
    features~\citep{tuli2022,xu2022,wu2023}. Moreover, there are applications
    where the underlying assumption of forecasting-based methods is questionable,
    \ie, when multiple legitimate forecast candidates exists. Think of an
    industrial machine switching between modes of operation upon unobserved
    conditions as in \citep{SRMH25}.
    Reconstruction-based methods, by contrast, learn a latent representation from which
    the input is regenerated, as in LSTM-based autoencoders~\citep{malhotra2015} and
    OmniAnomaly~\citep{su2019}. 
    However, such methods are prone to over-generalization:
    rather than failing to reconstruct anomalous inputs, the model generalises to unseen 
    anomalous patterns, thereby reducing sensitivity to genuine anomalies~\citep{song2023,kim2025}.
    More recently, graph-based methods have been adopted for \gls{mtsad}~\citep{deng2021} 
    to better capture spatial inter-variate dependencies. However, such methods 
    typically rely on a fixed graph structure, making them less effective in 
    scenarios where the underlying data distribution shifts over time~\citep{cai2026}.

    Progress in \gls{mtsad} is critically undermined by the widespread use of 
    ill-posed evaluation metrics~\citep{wu2023a}, under which random guessing 
    was shown to outperform state-of-the-art methods. Under more rigorous 
    evaluation protocols, deep learning methods remain competitive; however, 
    classical methods operating on spatial dependencies in state space prove 
    surprisingly strong, matching or surpassing far more complex approaches 
    on standard benchmarks~\citep{sarfraz2024}. This calls for classical 
    baselines to be treated as serious competitors.

    \textbf{Challenges in Industrial Time Series.}
    Industrial time series data is often sparse, irregularly sampled, and 
    noisy~\citep{colombi2024}. In practice, these challenges are commonly addressed 
    through preprocessing steps such as resampling onto regular grids, imputing missing 
    values, and applying denoising techniques~\citep{wang2025}. While these 
    strategies simplify downstream modeling, they inevitably introduce additional 
    bias into the data~\citep{belay2023}.
    
    Beyond preprocessing, the choice of model architecture also plays a critical 
    role in handling noise. Recent state-of-the-art approaches for multivariate 
    time series anomaly detection increasingly rely on transformer-based 
    architectures~\citep{xu2022,tuli2022}. However, transformers are known to be 
    prone to overfitting in the presence of noisy observations and often require 
    additional regularization or noise-robust training strategies to mitigate 
    this issue~\citep{kim2025}.    
    
    \textbf{Neural Differential Equations for Time Series Modeling.}
    A small but growing body of work leverages Neural ODEs~\citep{chen2018}, Neural 
    \glspl{sde}~\citep{li2020sde}, and, more broadly, Neural Differential 
    Equations~\citep{oh2025}, for anomaly detection in multivariate time 
    series~\citep{sun2026,glyndavies2022}. These models are, in principle, well-suited 
    for handling irregularly sampled and sparse observations, as their continuous-time 
    formulation naturally accommodates asynchronous measurements and reduces the need 
    for explicit resampling onto fixed grids.
    However, despite this apparent alignment, existing approaches rarely exploit these 
    properties explicitly in either their modeling design or evaluation protocols. In 
    practice, the advantages of continuous-time formulations are often underutilized, 
    with many methods effectively reverting to discretized settings.
    
    Neural \glspl{sde} are particularly promising in the presence of noisy 
    observations, as they explicitly model stochastic dynamics and capture uncertainty 
    in the underlying system. This enables a principled separation of intrinsic system 
    variability from observation noise, which is crucial for robust anomaly detection. 
    Yet, this capability remains largely unexplored in current anomaly detection 
    frameworks.

    \textbf{Positioning of This Work.} 
    Our proposed method, \gls{lsd}, distinguishes itself from existing
    \gls{mtsad} methods in two key aspects. First, to the best of our
    knowledge, \gls{lsd} is the first method to employ a latent \gls{sde} for
    \gls{mtsad}, modeling both temporal and spatial dependencies jointly in
    a low-dimensional latent manifold. By supporting both a Euclidean latent 
    space \Rn and a hyperspherical latent space \Sn, the latter being 
    particularly well-suited to roughly-cyclic data, \gls{lsd} provides a 
    geometrically informed representation absent from prior work.
    Second, unlike all existing methods, \gls{lsd} natively operates on sparse
    and irregularly sampled data by treating time as a continuous variable,
    eliminating the need for imputation or resampling as a preprocessing step.
    Furthermore, by employing an SDE rather than a Neural ODE, \gls{lsd}
    naturally accounts for noise in data.

    \section{Methodology} \label{sec:methodology} \ignore{ The challenge to the
        proposed method is \ldots restate goal: recon; MTSAD with Noise,
        irregularity, ir...    as a first-principle, we leverage the
        zugrundeliegende dynamic als inductiv bias latenten pfad
        representation, pfad im raum lernen, Kontinuierlich generierender
        Prozess, Nosie abbildet SDE, Shallow MLP, SDE zwing Dyn zu lernen Why
        SDE    
    }

    \subsection{Overview}

\begin{figure}[h]
        \centering
        \includegraphics[width=\linewidth]{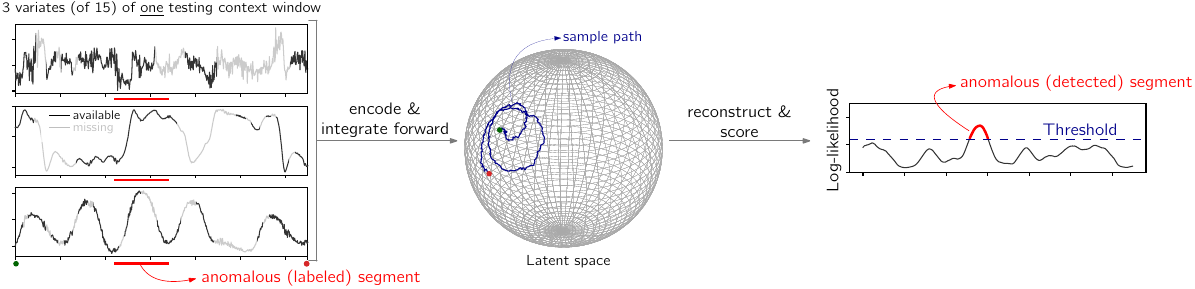}
        \caption{Overview of the proposed framework: a context window from the \gls{qad}
        dataset (three of 15 variates shown) is mapped during inference to a starting 
        point on the target space, here \Sn. From this point, the learned 
        \gls{sde} integrates forward to the desired evaluation time steps, and the 
        resulting reconstructions are scored to produce an anomaly metric.}
        \label{fig:ml-flow}
    \end{figure}

    We propose a reconstruction-based \gls{mtsad} algorithm that robustly detects 
    anomalies in noisy, sparse, and irregularly sampled multivariate time series. 
    As a guiding first principle, we exploit the 
    data's underlying latent dynamics as an inductive bias, modeling a continuous 
    path in a low-dimensional latent space. The observed time series is encoded 
    into an initial latent state, from which a generative process forecasts the 
    subsequent trajectory; the deviation between the reconstruction and the
    true observations serves as the anomaly score. The proposed approach is
    illustrated in \Cref{fig:ml-flow}.

    \glspl{sde} are a natural choice for these generative processes for three 
    reasons: 
    (a) irregular sampling is handled in continuous time without resampling, 
    (b) sparse inputs are naturally accommodated as the model does 
    not require observations across all features at every time step, and 
    (c) the stochastic component inherently accounts for observation noise.
    We therefore adopt the latent \gls{sde} model of \cite{zeng2023}, in which 
    latent variables evolve as trajectories on a low-dimensional homogeneous space.

    Unlike \cite{zeng2023}, we do not restrict ourselves to the spherical latent
    space \Sn and additionally study latent dynamics on a Euclidean space \Rn 
    with $\mathrm{GL}(n)$ action. 
    Furthermore, we exploit the log-likelihood of the reconstruction as an 
    anomaly score, which, to the best of our knowledge, has not been explored 
    in this context. 
    To prevent over-generalization to anomalous inputs, we employ a deliberately 
    weak decoder, which encourages the \gls{sde} to capture the true underlying 
    dynamics rather than overfit to individual observations.

    \ignore{
    \sthuinline{We need some glue text here from prior work to methodology.
        Plus, this is the perfect place where the intuition and the selling of the
        underlying intuition and the novelty should be praised. And for that we need
    a ML pipeline figure, too. This might be one of the critical parts of the paper, because the reviewer reads like this: introduction motivates and gives overview over contribution. prior work tells him we did our literature research and we have to explicitly name the research gaps. But then we have the big change in reading mode: What did the authors then do about the research gap, what shall I as a reviewer assess here.}
    }

    \subsection{Problem Formalization}
    \label{sec:problem-definition}

    We consider a $d$-variate, continuous time series as functions $\mathbf x:[0,T] \to \mathbb R^d$ with available observations at potentially unevenly spaced time points $t_i$ denoted as $\mathbf x_i = \mathbf x(t_i)$.     
    To discretize the time series, we compute the largest  value $\tau$ such that each point of time $t_k$ with available observations is a multiple of $\tau$.\footnote{For all considered datasets, this is indeed possible}
    We then represent $\mathbf x$ as an array $\mathbf{X} \in \mathbb R^{\frac{T}{\tau} \times d}$ together with a mask $\mathbf M \in \mathbb R^{\frac{T}{\tau} \times d}$ indicating at which multiples of $\tau$ observations are available.    
    Naturally, this formulation incorporates partial observations, \ie, it allows that not all $d$ features are available at an observation time $t_i$.

    In practice, a long time series $\mathbf x$ is split into $N$ windows corresponding to intervals of length $m\tau$, resulting in a dataset $\mathcal D = \{\mathbf X_1,\dots, \mathbf  X_N\}$ of arrays $\mathbf X_n \in \mathbb R^{m \times d}$ with accompanying masks $\mathbf M_n \in \mathbb R^{m \times d}$.
    An anomaly detection model $f_\Theta: \mathbb{R}^{d \times m} \rightarrow \mathbb{R}^m$ with parameters $\Theta$, mapping a window of $m$ observations across $d$ variates to a sequence of $m$ point-wise anomaly scores, is then trained on a training dataset $\mathcal{D}$ consisting \emph{only} of benign, non-anomalous, data in an unsupervised manner. The aim is then to detect anomalous data points within a test set $\mathcal{D}_\text{test}$.
    Specifically, the model produces a real-valued \emph{anomaly score} for each point of time, with larger values indicating anomalous behavior.
    This score is typically converted into binary predictions $\hat{\mathbf{y}}_n \in \{0,1\}^m$ via a threshold $r$, \ie, the predictions for a window $\mathbf X_n$ are given by $\hat{\mathbf{y}}_n = \mathbf{1}_{{f_\Theta(\mathbf X_n) > r}}$.

    Multi-trace datasets consist of $K$ independent traces
    $\mathcal{D} = \{\mathcal{D}_1, \ldots, \mathcal{D}_K\}$, where each trace is 
    acquired and treated 
    independently: a separate model is trained and evaluated on each 
    corresponding train/test pair, and the resulting metrics are aggregated 
    across all traces.

    \subsection{Modeling Normal Behavior via Latent SDEs}
    \label{sec:latent-sde-homogenous-space}
    
    To perform \gls{mtsad}, we leverage the framework of \cite{zeng2023}, modeling a time series $\mathbf x$ in terms of a latent time series $\mathbf z$ in a low dimensional latent space. This latent space is assumed a homogeneous space $H$ with some matrix Lie group $\mathcal G$ acting on it. The latent time series $\mathbf z$ is considered a sample path of a stochastic process $Z = G \cdot Z_0$, where $G$ again is a stochastic process in the Lie group $\mathcal G$ that acts on $H$. This process $G$ follows a certain SDE in $\mathcal G$, for details we refer to \cite[Section 3.2]{zeng2023}. 
    Importantly for our setting, the usage of an SDE naturally accounts for the inherent noise in the non-anomalous data that is present in many practical scenarios. Moreover, the quite specific form of $Z$ serves as a strong regularizer on the latent dynamics, that can be further controlled by the selection of the homogeneous space. In practice, we will use two forms of $Z$ given by the action of the rotation group $\mathrm{SO}(n)$ on a sphere $\mathbb S^n$ and by the action of the group $\mathrm{GL}(n)$ of invertible matrices on $\mathbb R^n$.

    %
    \textbf{Anomaly Detection.}  
    Once such a latent SDE model is trained, we can compute for each context window     $\mathbf{x}^i \in \mathbb{R}^{d \times w}$, the negative log-likelihood 
    $-\log p_\theta(\mathbf{x}^i \mid \mathbf{z})$ of the sample $\mathbf{x}^i$ under the learned latent path $\mathbf{z}$. 
    Optionally, the score is normalized via min-max scaling fitted on the training data. 

    \textbf{Implementation.} Full implementation details are provided in 
        Appendix~\ref{app:appendix-detailed-experimental-settings}.
    
    \section{Empirical Evaluation}
    \label{sec:empirical-evaluation}

    \textbf{Benchmark Datasets.} We evaluate \gls{lsd} and all baselines 
    on a diverse set of benchmarks spanning multiple domains and anomaly types: the 
    \gls{swat}~\citep{goh2017} and \gls{wadi}~\citep{chuadhry2017} datasets consist of 
    sensor readings and actuator states from a water treatment testbed, with 
    anomalies caused by cyber-physical attacks. The \gls{psm}~\citep{abdulaal2021} benchmark
    comprises multivariate telemetry from internal server infrastructure at eBay, 
    labelled for normal and anomalous behavior. \gls{msl} and 
    \gls{smap}~\citep{hundman2018} benchmarks contain telemetry from NASA's Mars Science 
    Laboratory rover (Curiosity) and the SMAP satellite mission, respectively, with 
    anomalies corresponding to reported telemetry faults. The \gls{smd}~\citep{su2019} 
    consists of server machine telemetry from a large internet company, with the 
    goal of detecting anomalous operational states.
    Finally, \gls{qad}~\citep{nosrati2026} is a recently introduced benchmark 
    of a distinct nature, comprising trajectories from a real cyber-physical system 
    that exhibit quasi-cyclic structure, \ie, cyclic movements in state space that 
    repeatedly return to approximately the same region without strict temporal 
    periodicity.
    
    \gls{swat}, \gls{wadi}, and \gls{psm} each consist of a single continuous 
    trace, whereas the remaining four benchmarks comprise multiple independent 
    traces; \gls{smd}, for instance, provides a separate train and test set per 
    server machine. Further details on all benchmarks, including dataset statistics, 
    are given in Appendix~\ref{app:appendix-benchmarks}.

    All datasets are normalized using min-max scaling. The rare missing values 
    in \gls{wadi} (less than $0.05\%$ of all observations) are imputed using 
    a centered moving average with a window of 10 samples.

    \textbf{Baselines.} We compare \gls{lsd} against 
    a comprehensive set of baselines encompassing both classical and deep 
    learning-based methods. Classical methods include \gls{pca}~\citep{shyu2006}, 
    \gls{knn}~\citep{ramaswamy2000}, \gls{ocsvm}~\citep{schoelkopf2001}, 
    \gls{lof}~\citep{he2003}, \gls{iforest}~\citep{liu2008}, and 
    \gls{copod}~\citep{li2020}.
    Deep learning-based methods 
    include DeepSVDD~\citep{ruff2018}, a deep one-class classification approach; 
    USAD~\citep{audibert2020} and TranAD~\citep{tuli2022}, both reconstruction-based 
    methods built on autoencoder and Transformer architectures, respectively; 
    TcnED~\citep{garg2022}, a temporal convolutional encoder-decoder; 
    AnomalyTransformer~\citep{xu2022}, which exploits association discrepancy in 
    the attention mechanism; DeepIF~\citep{xu2023}, a deep extension of Isolation 
    Forest; TimesNet~\citep{wu2023}, a general-purpose time series model adapted 
    for anomaly detection; and COUTA~\citep{xu2024}, a one-class classification 
    method with a Transformer backbone. All baselines are evaluated under the same 
    experimental protocol, using shared hyperparameter configurations where 
    applicable.
    

    \textbf{Metrics.} We evaluate all methods using three widely adopted metrics 
    for anomaly detection: the Area Under the Receiver Operating Characteristic 
    curve (AUROC), the Area Under the Precision-Recall Curve (AUPRC), and the 
    F1-score. AUROC and AUPRC are threshold-free and measure (anomaly detection) 
    performance across all operating points, making them particularly informative 
    for imbalanced datasets such as those encountered in anomaly detection. As the 
    F1-score requires a decision threshold, we follow \citep{sarfraz2024} and 
    select the threshold $r$ that maximises  the F1-score on the test data. 

    \subsection{Main Results}\label{sec:comparison-baselines}
    Both 
    variants, \gls{lsd} with \Rn and \Sn as latent manifold, incorporate a data subsampling step during training, which randomly masks a
    subset of input samples at each iteration, acting as a regularizer that mitigates overfitting
    of the learned \gls{sde}. Since the mask is sampled independently at every update step during stochastic optimization, the model is exposed to the complete dataset over the course of training.

    Experimental results for all baselines and the proposed \gls{lsd}, trained in both the
    hyperspherical space $\mathbb{S}^n$ and the Euclidean space $\mathbb{R}^n$, are reported in
    \Cref{tab:results_benchmarks_single_trace,tab:results_benchmarks_multi_trace}.
    The tables are organized in three parts: classical machine learning methods, state-of-the-art
    methods, and the proposed \gls{lsd} in both $\mathbb{S}^n$ and $\mathbb{R}^n$.
    \Cref{tab:results_benchmarks_single_trace} reports results for \gls{swat}, \gls{wadi}, and
    \gls{psm}, each consisting of a single data trace, while
    \Cref{tab:results_benchmarks_multi_trace} covers \gls{smap}, \gls{msl}, and \gls{smd},
    where the number of traces is indicated in brackets.
    
     \begin{table}
         \caption{Performance comparison on single-trace benchmarks. The 
         \colorbox{first}{best}, \colorbox{second}{second-best}, and 
         \colorbox{third}{third-best} results per metric are highlighted.
         Within each section, methods are ranked in decreasing order of mean rank.}
         \label{tab:results_benchmarks_single_trace}
         \centering
         \setlength{\tabcolsep}{3pt} 
         \scriptsize 
         \resizebox{\textwidth}{!}{
         \begin{tabular}{cl ccc ccc ccc}
             \toprule
             && \multicolumn{3}{c}{\small \textbf{SWaT}} & \multicolumn{3}{c}{\small \textbf{WaDi}} & \multicolumn{3}{c}{\small \textbf{PSM}} \\
             \cmidrule(lr){3-5}\cmidrule(lr){6-8}\cmidrule(lr){9-11}
             \textbf{Rank}\big\downarrow & \textbf{Model}& \textbf{AUC} \big\uparrow & \textbf{AUPRC} \big\uparrow & \textbf{F1} \big\uparrow & \textbf{AUC} \big\uparrow & \textbf{AUPRC} \big\uparrow & \textbf{F1} \big\uparrow & \textbf{AUC} \big\uparrow & \textbf{AUPRC} \big\uparrow & \textbf{F1} \big\uparrow \\
             
             \midrule

             \rankbox[second]{5.56} & COPOD & \cellcolor{first}{87.04 \std{0.00}} & \cellcolor{first}{76.59 \std{0.00}} & 75.22 \std{0.00} & \cellcolor{first}{78.78 \std{0.00}} & \cellcolor{second}{22.22 \std{0.00}} & \cellcolor{first}{35.98 \std{0.00}} & 66.87 \std{0.00} & 43.24 \std{0.00} & 49.01 \std{0.00} \\
             \rankbox[third]{5.56} & IForest & \cellcolor{second}{84.93 \std{0.51}} & \cellcolor{third}{73.76 \std{0.63}} & 74.49 \std{0.01} & \cellcolor{second}{73.90 \std{0.52}} & 17.53 \std{1.07} & \cellcolor{second}{29.63 \std{2.10}} & 70.01 \std{1.18} & 47.06 \std{1.61} & 50.59 \std{1.13} \\
             \rankbox{6.33} & OCSVM & 82.76 \std{0.00} & 72.65 \std{0.00} & \cellcolor{second}{77.45 \std{0.00}} & 52.27 \std{0.00} & \cellcolor{first}{27.69 \std{0.00}} & 13.04 \std{0.00} & 69.83 \std{0.00} & 51.29 \std{0.00} & 49.63 \std{0.00} \\
             \rankbox{7.22} & KNN & 82.54 \std{0.00} & 72.49 \std{0.00} & \cellcolor{first}{78.83 \std{0.00}} & 47.10 \std{0.00} & 5.00 \std{0.00} & 12.62 \std{0.00} & 73.91 \std{0.00} & \cellcolor{first}{54.33 \std{0.00}} & {54.86 \std{0.00}} \\
             \rankbox{8.11} & PCA & 82.99 \std{0.00} & 73.34 \std{0.00} & 76.07 \std{0.00} & 49.95 \std{0.00} & 5.21 \std{0.00} & 13.53 \std{0.00} & 65.74 \std{0.00} & 47.44 \std{0.00} & 46.59 \std{0.00} \\ 
             \rankbox{11.78} & LOF & 48.20 \std{0.00} & 15.63 \std{0.00} & 25.31 \std{0.00} & 46.66 \std{0.00} & 4.94 \std{0.00} & 11.93 \std{0.00} & 71.57 \std{0.00} & 43.61 \std{0.00} & 54.48 \std{0.00} \\
             
             \midrule
             
             \rankbox{7.89} &TcnED & \cellcolor{third}{}{83.91 \std{0.22}} & \cellcolor{second}{74.43 \std{0.23}} & {76.50 \std{0.30}} & {53.14 \std{0.73}} & 11.04 \std{1.50} & 14.14 \std{1.67} & 61.56 \std{0.31} & 37.51 \std{0.28} & 46.05 \std{0.40} \\
             \rankbox{8.78} &TranAD & 83.07 \std{0.25} & 73.29 \std{0.17} & 76.07 \std{0.14} & 49.13 \std{0.11} & 5.14 \std{0.01} & 12.67 \std{0.08} & 66.34 \std{0.47} & 46.86 \std{0.73} & 48.43 \std{0.36} \\
             \rankbox{8.78} &AnomalyTrans. & 82.82 \std{0.21} & 72.67 \std{0.32} & 76.08 \std{0.34} & 46.36 \std{0.59} & 4.85 \std{0.04} & 12.84 \std{0.09} & 70.04 \std{1.44} & 48.50 \std{1.07} & 50.82 \std{1.75} \\
             \rankbox{9.56} & USAD & 82.85 \std{0.12} & 73.38 \std{0.06} & \cellcolor{third}{76.81 \std{0.01}} & 48.99 \std{0.16} & 5.06 \std{0.01} & 12.96 \std{0.26} & 61.78 \std{2.14} & 40.18 \std{2.26} & 46.32 \std{1.14} \\
             \rankbox{10.00} &DeepIF & 82.94 \std{0.35} & 72.01 \std{0.48} & 75.30 \std{0.43} & 48.04 \std{0.94} & 5.00 \std{0.09} & 12.60 \std{0.25} & 69.47 \std{0.67} & 46.11 \std{0.75} & 49.87 \std{0.86} \\
             \rankbox{10.11} &COUTA & 78.30 \std{10.09} & 50.85 \std{25.57} & 59.54 \std{18.64} & 44.46 \std{3.67} & {6.90 \std{4.46}} & 13.30 \std{3.47} & 70.72 \std{2.72} & 45.98 \std{4.17} & 51.92 \std{2.54} \\
             \rankbox{11.22} & DeepSVDD & 46.79 \std{2.08} & 13.58 \std{0.99} & 27.75 \std{0.89} & 39.90 \std{1.43} & 4.40 \std{0.09} & 11.07 \std{0.12} & \cellcolor{second}{77.03 \std{2.01}} & 50.74 \std{3.54} & \cellcolor{second}{57.99 \std{3.49}} \\
             \rankbox{13.89} &TimesNet & 24.45 \std{0.54} & {8.72 \std{0.09}} & 21.38 \std{0.00} & 50.47 \std{0.36} & {6.04 \std{0.09}} & 11.72 \std{0.39} & 59.58 \std{0.15} & 39.24 \std{0.31} & 43.54 \std{0.05} \\

             \midrule

             \rankbox[first]{5.44} &LSD on $\mathbb{R}^n$ (ours) & {82.26 \std{0.05}} & 70.80 \std{0.05} & 74.86 \std{0.06} & {54.84 \std{0.25}} & 20.63 \std{0.03} & 26.14 \std{0.03}& \cellcolor{first}{77.23 \std{1.56}} & \cellcolor{second}{53.92 \std{1.17}} &\cellcolor{first}{59.87 \std{1.85}} \\
             \rankbox{5.78} &LSD on $\mathbb{S}^n$ (ours) & 82.26 \std{0.05} & 70.80 \std{0.05} & 74.86 \std{0.06}& \cellcolor{third}{54.95 \std{0.14}} & \cellcolor{third}{20.64 \std{0.01}} & \cellcolor{third}{26.14 \std{0.00}} & \cellcolor{third}{75.64 \std{1.29}} & \cellcolor{third}{52.37 \std{2.04}} & \cellcolor{third}{56.13 \std{1.77}} \\
            
             \bottomrule
         \end{tabular}
         }
    \end{table}
    
    \begin{table}
        \caption{Performance comparison on multi-trace benchmarks. The 
            \colorbox{first}{best}, \colorbox{second}{second-best}, and 
            \colorbox{third}{third-best} results per metric are highlighted.
            Within each section, methods are ranked in decreasing order of mean rank.
            }
        \label{tab:results_benchmarks_multi_trace}
        \centering
        \setlength{\tabcolsep}{3pt} 
        \scriptsize 

        \resizebox{\textwidth}{!}{
        \begin{tabular}{cl ccc ccc ccc}
            \toprule
             && \multicolumn{3}{c}{\small \textbf{SMAP (55)}} & \multicolumn{3}{c}{\small \textbf{MSL (27)}} & \multicolumn{3}{c}{\small \textbf{SMD (28)}} \\
             \cmidrule(lr){3-5}\cmidrule(lr){6-8}\cmidrule(lr){9-11}
             \textbf{Rank}\big\downarrow & \textbf{Model}& \textbf{AUC} \big\uparrow & \textbf{AUPRC} \big\uparrow & \textbf{F1} \big\uparrow & \textbf{AUC} \big\uparrow & \textbf{AUPRC} \big\uparrow & \textbf{F1} \big\uparrow & \textbf{AUC} \big\uparrow & \textbf{AUPRC} \big\uparrow & \textbf{F1} \big\uparrow \\
            
            \midrule
            
            \rankbox{5.89} & KNN & 61.70 \std{0.00} & 26.71 \std{0.00} & 36.48 \std{0.00} & \cellcolor{third}{67.02 \std{0.00}} & {26.43 \std{0.00}} & {39.14 \std{0.00}} & 81.80 \std{0.00} & 43.13 \std{0.00} & 48.99 \std{0.00} \\
            \rankbox{6.89}& PCA & \cellcolor{second}{64.89 \std{0.00}} & 27.18 \std{0.00} & {37.39 \std{0.00}} & 63.28 \std{0.00} & 20.70 \std{0.00} & 34.06 \std{0.00} & 80.61 \std{0.00} & 43.55 \std{0.00} & 48.98 \std{0.00} \\
            \rankbox{9.22}& OCSVM & 63.87 \std{0.00} & {27.13 \std{0.00}} & 34.99 \std{0.00} & 60.62 \std{0.00} & {24.18 \std{0.00}} & 31.87 \std{0.00} & 79.21 \std{0.00} & 41.63 \std{0.00} & 40.00 \std{0.00} \\
            \rankbox{12.00}& COPOD & 62.62 \std{0.00} & 20.28 \std{0.00} & 33.12 \std{0.00} & 63.74 \std{0.00} & 20.24 \std{0.00} & 33.37 \std{0.00} & 77.70 \std{0.00} & 29.57 \std{0.00} & 33.48 \std{0.00} \\
            \rankbox{12.11}& LOF & 58.32 \std{0.00} & 25.67 \std{0.00} & 33.77 \std{0.00} & 61.48 \std{0.00} & 20.74 \std{0.00} & 32.79 \std{0.00} & 71.20 \std{0.00} & 30.71 \std{0.00} & 36.73 \std{0.00} \\
            \rankbox{15.44}& IForest & 57.86 \std{0.21} & 19.55 \std{0.17} & 29.33 \std{0.12} & 57.39 \std{0.07} & 17.45 \std{0.14} & 27.06 \std{0.28} & 77.43 \std{0.15} & 27.12 \std{0.44} & 32.43 \std{0.35} \\           
            
            \midrule

            \rankbox[third]{5.11}&TcnED & \cellcolor{first}{65.35 \std{0.48}} & {26.76 \std{0.43}} & \cellcolor{third}{37.46 \std{0.50}} & 63.77 \std{0.38} & 21.20 \std{0.09} & 34.44 \std{0.29} & 81.33 \std{0.17} & \cellcolor{first}{45.36 \std{0.31}} & \cellcolor{first}{49.72 \std{0.28}} \\
            \rankbox{5.22}&USAD & 59.90 \std{0.29} & 26.05 \std{0.15} & 37.22 \std{0.19} & \cellcolor{first}{69.51 \std{0.01}} & \cellcolor{third}{28.67 \std{0.08}} & \cellcolor{third}{39.86 \std{0.03}} & \cellcolor{third}{86.11 \std{0.27}} & 37.93 \std{0.22} & \cellcolor{second}{49.49 \std{0.39}} \\
            \rankbox{5.89} &DeepIF & 63.13 \std{1.29} & \cellcolor{second}{29.68 \std{0.59}} & \cellcolor{first}{40.08 \std{0.73}} & 61.73 \std{0.40} & 22.32 \std{0.84} & 35.14 \std{0.75} & \cellcolor{first}{87.58 \std{0.16}} & 37.66 \std{0.21} & 48.66 \std{0.25} \\
            \rankbox{6.67}&TranAD & 62.03 \std{1.82} & 24.76 \std{0.39} & 35.29 \std{0.80} & 64.05 \std{1.21} & 21.39 \std{0.42} & 33.75 \std{0.39} & 82.92 \std{0.41} & \cellcolor{second}{44.46 \std{0.18}} & \cellcolor{third}{49.47 \std{0.27}} \\
            \rankbox{8.78} &TimesNet & 55.12 \std{0.98} & 21.21 \std{1.03} & 33.00 \std{0.80} & {65.24 \std{0.63}} & 24.84 \std{0.27} & 36.27 \std{0.66} & \cellcolor{second}{86.34 \std{0.16}} & 39.37 \std{0.63} & 45.29 \std{0.47} \\
            \rankbox{9.44} &COUTA & 59.68 \std{5.67} & 24.14 \std{0.76} & 35.06 \std{0.42} & 63.52 \std{1.21} & 20.56 \std{0.84} & 31.83 \std{0.8} & 82.07 \std{0.49} & \cellcolor{third}{43.78 \std{0.40}} & 48.92 \std{0.47} \\
            \rankbox{10.33}&DeepSVDD & \cellcolor{third}{64.78 \std{1.11}} & 22.12 \std{1.07} & 35.45 \std{0.68} & 64.05 \std{1.31} & 20.06 \std{1.05} & 31.19 \std{1.46} & 82.52 \std{0.68} & 25.32 \std{2.01} & 35.82 \std{1.89} \\            
            \rankbox{14.22}&AnomalyTrans. & 57.80 \std{3.88} & 20.49 \std{1.84} & 31.51 \std{1.95} & 57.65 \std{3.68} & 17.64 \std{1.31} & 26.99 \std{1.86} & 80.08 \std{0.33} & 28.96 \std{0.96} & 38.53 \std{0.45} \\
            
            \midrule

            \rankbox[first]{4.00} &LSD on $\mathbb{R}^n$ (ours) & {62.55 \std{0.37}} & \cellcolor{first}{29.97 \std{0.48}} & \cellcolor{second}{38.52 \std{0.29}} &\cellcolor{second}{67.10 \std{0.46}} & \cellcolor{first}{30.62 \std{0.92}} &\cellcolor{first}{42.11 \std{0.51}}& 82.74 \std{0.37} & 40.94 \std{0.54} &46.85 \std{0.53} \\
            \rankbox[second]{4.78} &LSD on $\mathbb{S}^n$ (ours) & 61.25 \std{1.30} & \cellcolor{third}{29.06 \std{0.69}} & {37.45 \std{0.25}} & {66.31 \std{1.63}} & \cellcolor{second}{28.93 \std{1.24}} & \cellcolor{second}{40.28 \std{1.15}} & 81.89 \std{0.56} & 42.31 \std{1.08} & 49.00 \std{0.67} \\
            
            \bottomrule
        \end{tabular}
        }
    \end{table}


    To compare methods across benchmarks, we rank each metric within the table and report
    the mean rank over all metrics and benchmarks, giving a measure of overall method
    performance. The ranks displayed alongside each method's name in
    \Cref{tab:results_benchmarks_single_trace,tab:results_benchmarks_multi_trace}
    are computed over the benchmark of the corresponding table, the joint ranking over all 
    six benchmarks is shown in \Cref{tab:rankings}.
    
    The rank scores clearly demonstrate the overall advantage of \gls{lsd} across 
    all benchmark configurations. On the single-trace benchmarks, \gls{lsd} in 
    \Rn achieves the best overall rank, with the \Sn variant 
    ranking fourth. On the multi-trace benchmarks (\Cref{tab:results_benchmarks_multi_trace}), 
    both variants outperform all baselines, with the \Rn variant 
    achieving a mean rank improvement of $\approx$ 1.22 over the next best 
    baseline. 
    
    A noteworthy observation is that the two benchmark types reveal complementary 
    strengths among the baselines: classical machine learning methods perform 
    prominently on single-trace benchmarks, whereas deep learning-based methods 
    show a relative advantage on multi-trace benchmarks. Nevertheless, 
    as illustrated in \Cref{tab:rankings}, the proposed \gls{lsd} 
    achieves a competitive advantage over all baselines consistently across both 
    settings, demonstrating its robustness to benchmark characteristics.

    \FloatBarrier
    \begin{wraptable}{R}{0.45\textwidth}
        \vspace{-0.45cm}
        \caption{Method rankings sorted by overall rank, reported over all benchmarks combined, as well as separately over the single-trace and multi-trace benchmarks.}
        \label{tab:rankings}
        \centering
        \setlength{\tabcolsep}{3pt} 
        \scriptsize 
        
        \begin{tabular}{lrrr}
            \toprule
            \textbf{Model} & \textbf{Overall} & \textbf{Single Trace} & \textbf{Multi Trace} \\
            \midrule
            LSD on \Rn (ours) & \cellcolor{first}{4.72} & \cellcolor{first}{5.44} & \cellcolor{first}{4.00} \\
            LSD on \Sn (ours) & \cellcolor{second}{5.28} & 5.78 & \cellcolor{second}{4.78} \\
            TcnED & \cellcolor{third}{6.50} & 7.89 & \cellcolor{third}{5.11} \\
            KNN & 6.56 & 7.22 & 5.89 \\
            USAD & 7.39 & 9.56 & 5.22 \\
            PCA & 7.50 & 8.11 & 6.89 \\
            TranAD & 7.72 & 8.78 & 6.67 \\
            OCSVM & 7.78 & 6.33 & 9.22 \\
            DeepIF & 7.94 & 10.00 & 5.89 \\
            COPOD & 8.78 & \cellcolor{second}{5.56} & 12.00 \\
            COUTA & 9.78 & 10.11 & 9.44 \\
            IForest & 10.50 & \cellcolor{third}{5.56} & 15.44 \\
            DeepSVDD & 10.78 & 11.22 & 10.33 \\
            TimesNet & 11.33 & 13.89 & 8.78 \\
            AnomalyTrans. & 11.50 & 8.78 & 14.22 \\
            LOF & 11.94 & 11.78 & 12.11 \\
            \bottomrule
        \end{tabular}
    \end{wraptable}

    The effectiveness of the proposed method on real-world industrial data is demonstrated on the
    \gls{qad}~\citep{nosrati2026} benchmark. Unlike the other benchmarks, information about the
    underlying dynamic is revealed by data analysis:
    the system performs a quasi-cyclic pick-and-place robotic task, returning to the
    same initial state after each iteration. 
    Each of the 16 traces is first subsampled evenly to \SI{10}{\percent} of its original length, resulting
    in a \SI{10}{\hertz} sampling rate.
    Leveraging knowledge of the period length, we
    hypothesize that exposing \gls{lsd} to roughly one \emph{full} cycle allows it to capture
    richer temporal structure: the latent space of the \gls{lsd}, especially the compact \Sn variant, 
    introduces an inductive bias that enables explicit modeling of the repetitive dynamics on the lower 
    dimensional hypersphere \Sn.
    We therefore set the context window length to 500 with a stride of
    500, yielding non-overlapping windows, which approximates the average period length of the repetitive task.
    The results for the \gls{qad} benchmark are reported
    in \Cref{tab:results_benchmark_qad}.

    The results reveal a surprising finding: the best-performing method on the \gls{qad} benchmark
    across all three metrics is \gls{lof}, which ranked by far the worst on the other six benchmarks.
    \gls{knn} consistently ranks second, followed by DeepIF. Both \gls{lsd} variants rank fourth and
    fifth, at nearly half the AUPRC of \gls{lof}. Notably, the remaining
    state-of-the-art methods rank comparatively low, in stark contrast to their strong performance on
    the benchmarks in \Cref{tab:results_benchmarks_single_trace,tab:results_benchmarks_multi_trace}.
    
    \FloatBarrier
    \begin{wraptable}{R}{0.55\textwidth}
        \vspace{-0.445cm}
        \caption{Performance comparison on the \gls{qad} benchmark. The 
            \colorbox{first}{best}, \colorbox{second}{second-best}, and 
            \colorbox{third}{third-best} results per metric are highlighted.
            Within each section, methods are ranked in decreasing order of mean rank.}
        \label{tab:results_benchmark_qad}
        \centering
        \setlength{\tabcolsep}{3pt} 
        \scriptsize 
        
        \begin{tabular}{cl ccc}
            \toprule
             && \multicolumn{3}{c}{\small \textbf{QAPPD (16)}} \\
             \cmidrule(lr){3-5}
            \textbf{Rank}\big\downarrow &\textbf{Model} & \textbf{AUC} \big\uparrow & \textbf{AUPRC} \big\uparrow & \textbf{F1} \big\uparrow\\
            
            \midrule

            \rankbox[first]{1.0} & LOF & \cellcolor{first}{77.11 \std{0.00}} & \cellcolor{first}{17.83 \std{0.00}} & \cellcolor{first}{27.73 \std{0.00}} \\
            \rankbox[second]{2.0} & KNN & \cellcolor{second}{67.97 \std{0.00}} & \cellcolor{second}{16.31 \std{0.00}} & \cellcolor{second}{23.06 \std{0.00}} \\
            \rankbox{7.7} & IForest & 60.67 \std{2.15} & 7.83 \std{0.51} & 13.62 \std{0.40} \\
            \rankbox{10.0} & COPOD & 60.78 \std{0.00} & 8.79 \std{0.00} & 14.97 \std{0.00} \\
            \rankbox{10.7} & OCSVM & 59.26 \std{0.00} & 6.70 \std{0.00} & 12.68 \std{0.00} \\
            \rankbox{11.3} & PCA & 57.76 \std{0.00} & 6.48 \std{0.00} & 13.05 \std{0.00} \\
            
            \midrule
            
            \rankbox[third]{3.3} & DeepIF & 64.06 \std{2.85} & \cellcolor{third}{12.31 \std{0.77}} & \cellcolor{third}{18.73 \std{1.10}} \\
            \rankbox{9.3} & USAD & 52.48 \std{0.23} & 9.69 \std{0.14} & 15.57 \std{0.08} \\
            \rankbox{9.7} & TcnED & 57.08 \std{2.45} & 7.48 \std{0.37} & 13.21 \std{0.34} \\
            \rankbox{9.7} & COUTA & 61.56 \std{3.51} & 6.52 \std{0.70} & 12.77 \std{1.65} \\
            \rankbox{11.0} & TranAD & 55.92 \std{1.46} & 6.84 \std{0.30} & 13.07 \std{0.20} \\
            \rankbox{12.3} & DeepSVDD & 59.17 \std{1.45} & 6.33 \std{0.14} & 11.90 \std{0.23} \\            
            \rankbox{14.0} & TimesNet & 56.63 \std{2.92} & 6.32 \std{0.21} & 11.86 \std{0.49} \\ 
            \rankbox{15.3} & AnomalyTrans. & 54.48 \std{3.52} & 6.22 \std{0.54} & 11.56 \std{1.03} \\
            
            \midrule
            
            \rankbox{4.0} & LSD on $\mathbb{S}^n$ (ours) & \cellcolor{third}{64.46 \std{0.78}} & {10.16 \std{0.36}} & {16.97 \std{0.61}} \\
            \rankbox{4.7} & LSD on $\mathbb{R}^n$ (ours) & 64.05 \std{0.74} & 10.01 \std{0.41} & 17.58 \std{0.60} \\
            
            \bottomrule
        \end{tabular}
    \end{wraptable}

    \newpage    
    \subsection{Sparsity Evaluation}\label{sec:sparsity-evaluation}
    To evaluate the robustness of the proposed method under varying levels of data sparsity, we compare 
    against the three best-performing baselines, selected by mean rank from \Cref{tab:results_benchmark_qad}. All runs share a fixed hyperparameter 
    configuration; only the masking ratio is varied.
    
    To simulate data sparsity and the consequent irregular sampling, we generate masks
    that specify the available data points as follows.
    For each feature, we sample lengths of consecutive zeros, \ie, of missing observations, in the mask from a geometric distribution with parameter $s p$ and lengths of consecutive ones from a geometric distribution with parameter $s(1-p)$. 
    This corresponds to a discrete Markov chain with probabilities 
    $\mathbb P[X_{t+1}=1 | X_t=0] = s p$ and  $\mathbb P[X_{t+1}=0 | X_t=1] = s (1-p)$ and creates a mask with expected average $p$.  
    Compared to a uniformly sampled random mask, this results in longer intervals of missing or available information, and therefore a more difficult problem, but also a realistic setting. The lengths of such intervals is further controlled by parameter $s\in (0,1)$ (in the experiment we use $s=0.2$) with smaller $s$ resulting in more long intervals. 
    The mask is applied to the entire time series, \ie before dividing it into windows.
    Since the baselines cannot natively handle missing data, masked intervals are imputed 
    by linear interpolations between the neighboring unmasked observations. 
    Masking is applied uniformly across the training, validation, and test splits.
    
    \begin{figure}
        \centering
        \includegraphics[width=1.0\linewidth]{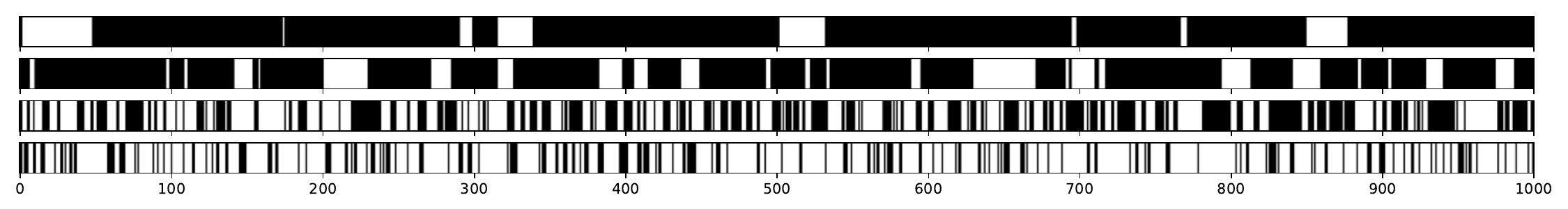}
        \caption{Exemplary masks resulting from differing sparsity parameters $p$ ($0.1, 0.2, 0.5, 0.75$ from top to bottom). Black indicates missing observations.}
        \label{fig:masks}
        \vspace{-0.3cm}
    \end{figure}

    All baselines, unlike the proposed \gls{lsd}, are unable to natively handle 
    sparse and irregularly sampled inputs. We therefore apply two standard 
    preprocessing steps: the data is \emph{resampled} to the original acquisition 
    frequency, and missing values on each variate are imputed via linear 
    interpolation between the nearest preceding and following observations. Values 
    at the beginning or end of a sequence are filled by forward or backward 
    carrying the first or last available observation, respectively.
    
    We report results for \emph{sparse burst subsampling} at \SI{1}{\percent} and \SI{5}{\percent} 
    masking ratios. \gls{lsd} is compared against the top-3 ranked baselines from 
    \Cref{tab:results_benchmark_qad}, with results averaged over 5 runs reported 
    in \Cref{tab:results_sparsity_qad}.
    The results clearly demonstrate that all three baselines (\gls{knn}, 
    \gls{lof}, and DeepIF) fail to compensate for sparsity even when linear 
    interpolation as preprocessing is applied.
    \gls{knn} degrades by $44.76\%$ and $45.68\%$ in AUPRC 
    at \SI{1}{\percent} and \SI{5}{\percent} subsampling, respectively. The impact is even more 
    pronounced for \gls{lof}, which suffers reductions of $63.77\%$ and $67.02\%$, 
    and for DeepIF, with degradations of $60.28\%$ and $37.94\%$ at $1\%$ and 
    $5\%$, respectively.
    In contrast, \gls{lsd} maintains a stable performance across both 
    subsampling ratios. On $\mathbb{R}^n$, \gls{lsd} incurs a drop of only $3.2\%$ 
    in AUPRC and $4.89\%$ in F1, more than an order of magnitude smaller than 
    the degradations observed for the baselines.

    \begin{table}
        \caption{Sparsity evaluation on the \gls{qad} benchmark. Three configurations are shown, 
        each indicated with the share of original data available in parentheses.}
        \label{tab:results_sparsity_qad}
        \centering
        \setlength{\tabcolsep}{3pt} 
        \scriptsize 

        \resizebox{\textwidth}{!}{
        \begin{tabular}{cl ccc ccc ccc}
            \toprule
             && \multicolumn{3}{c}{\small \textbf{QAPPD (1\%)}} & \multicolumn{3}{c}{\small \textbf{QAPPD (5\%)}} & \multicolumn{3}{c}{\small \textbf{QAPPD (100\%)}} \\
             \cmidrule(lr){3-5}\cmidrule(lr){6-8}\cmidrule(lr){9-11}
            \textbf{Rank}\big\downarrow & \textbf{Model} & \textbf{AUC} \big\uparrow & \textbf{AUPRC} \big\uparrow & \textbf{F1} \big\uparrow & \textbf{AUC} \big\uparrow & \textbf{AUPRC} \big\uparrow & \textbf{F1} \big\uparrow & \textbf{AUC} \big\uparrow & \textbf{AUPRC} \big\uparrow & \textbf{F1} \big\uparrow \\
            
            \midrule
            
            \rankbox[second]{2.78} & KNN & \cellcolor{third}{57.61 \std{2.87}}  & \cellcolor{third}{9.01 \std{1.80}} & \cellcolor{third}{15.85 \std{1.69}}& {56.82 \std{2.43}}  & \cellcolor{third}{8.86 \std{1.44}} & \cellcolor{third}{14.79 \std{1.48}} & \cellcolor{second}{67.97 \std{0.00}} & \cellcolor{second}{16.31 \std{0.00}} & \cellcolor{second}{23.06 \std{0.00}} \\
            
            \rankbox{3.33} & LOF & {54.89 \std{1.62}}  & {6.46 \std{0.67}} & {13.84 \std{1.20}} & {52.58 \std{0.63}}  & {5.88 \std{0.19}} & {11.50 \std{0.39}} & \cellcolor{first}{77.11 \std{0.00}} & \cellcolor{first}{17.83 \std{0.00}} & \cellcolor{first}{27.73 \std{0.00}} \\
            
            \rankbox{4.00} & DeepIF & 46.46 \std{3.04} & 4.89 \std{0.78} & 11.74 \std{1.39} & \cellcolor{third}{60.16 \std{1.97}} & 7.64 \std{0.57} & 14.49 \std{1.13} & 64.06 \std{2.85} & \cellcolor{third}{12.31 \std{0.77}} & \cellcolor{third}{18.73 \std{1.10}} \\
            
            \rankbox[first]{2.11} & LSD on $\mathbb{S}^n$ (ours) & \cellcolor{first}{64.50 \std{0.17}} & \cellcolor{first}{9.94 \std{0.66}} & \cellcolor{first}{16.81 \std{0.97}} & \cellcolor{first}{64.85 \std{0.96}} & \cellcolor{second}{9.83 \std{0.38}} & \cellcolor{first}{16.80 \std{0.65}} & \cellcolor{third}{64.46 \std{0.78}} & {10.16 \std{0.36}} & {16.97 \std{0.61}} \\
            
            \rankbox[second]{2.78} & LSD on $\mathbb{R}^n$ (ours) & \cellcolor{second}{64.06 \std{1.15}} & \cellcolor{second}{9.69 \std{0.47}} & \cellcolor{second}{16.72 \std{0.91}} & \cellcolor{second}{64.35 \std{0.14}}& \cellcolor{first}{9.93 \std{0.47}} & \cellcolor{second}{16.61 \std{0.68}} & 64.05 \std{0.74} & 10.01 \std{0.41} & 17.58 \std{0.60} \\
            \bottomrule
        \end{tabular}
        }
    \end{table}  

    \textbf{Runtime measurements.} A detailed runtime comparison of \gls{lsd}
    against all baseline methods is provided in Appendix~\ref{app:appendix-runtimes}.

    \section{Discussion and Limitations}
    \label{sec:discussion-limitations}
    In this paper, we have presented \gls{lsd}, an \gls{sde}-based modeling framework
    for \gls{mtsad} in a latent homogenous space, either, but not limited to, the Euclidean \Rn or 
    the spherical \Sn.
    The proposed method specifically embeds the test data in the learned space and
    the likelihood of the reconstruction by the decoder is utilized as an anomaly score.
    Although the used framework would also allow for forecasting or to be used as a generative model by itself, we do not explicitly 
    utilize this.
    \gls{lsd} yields consistently strong performance across all evaluated 
    benchmarks, with a more pronounced advantage over the baselines on 
    multi-trace benchmarks than on single-trace benchmarks.
    We further demonstrate that \gls{lsd} does not suffer significant performance 
    degradation under bursty subsampling, a sparsity pattern commonly encountered 
    in industrial control systems, while all baseline methods degrade substantially.
    
    Our results on the \gls{qad} benchmark highlight the favorable inductive bias 
    introduced by the choice of latent space geometry. In particular, the 
    hyperspherical latent space \Sn consistently outperforms its 
    Euclidean counterpart \Rn, a trend that contrasts with results on 
    the remaining benchmarks. This suggests that modeling cyclic dynamics on 
    the hypersphere provides meaningful structural advantages, as the geometry 
    naturally aligns with the underlying periodicity of the data. Furthermore, the 
    results on this benchmark underscore the need for evaluation protocols that 
    explicitly account for cyclic characteristics: none of the baseline methods 
    achieve meaningful performance, exposing a significant gap in the capability 
    of current state-of-the-art approaches on this problem class.

    \textbf{Limitations.} The primary limitation of \gls{lsd} relative to all 
    baseline methods is its  substantially higher training cost, inherent to 
    the complexity of fitting a latent \gls{sde} with a numerical solver.
    Additionally, the sparsity experiment is constrained by the use of artificially 
    introduced ``missingness'', which may not fully reflect the irregular and 
    heterogeneous sparsity patterns encountered in real industrial deployments. 
    Finally, unlike some prior works, we deliberately refrained from extensive 
    hyperparameter tuning, including the selection of regularization strategies, 
    score and data normalization techniques, and likelihood aggregation methods, 
    leaving open the possibility that further tuning could yield additional 
    performance gains.
    
    \textbf{Future Work.} The sparsity experiments on \gls{qad} highlight the 
    need for benchmarks that exhibit cyclic dynamics alongside natural 
    sparsity and true irregular sampling, i.e., characteristics that are prevalent in 
    real industrial deployments but remain underrepresented in current \gls{mtsad} 
    benchmarks. Developing and releasing such benchmarks represents an important 
    direction for future work. Beyond benchmarking, the \gls{lsd} framework is 
    a natural candidate for extension to related tasks such as \emph{root cause 
    analysis}, where the per-variate log-likelihood may serve as a principled 
    indicator for isolating the source of anomalous behavior.
    \ignore{quasi periodic \textrightarrow torus \textrightarrow propose a novel framework}
    Finally, following the guiding first principle of geometry-informed latent 
    spaces, the framework could be extended to support an $n$-dimensional torus 
    $\mathbb{T}^n = \mathbb{S}^1 \times \cdots \times \mathbb{S}^1$ as the latent 
    manifold, which may be particularly well-suited for modeling strictly
    periodic ($n=1$) or quasi-periodic ($n>1$) data, \ie, possessing multiple periods with
    incommensurate lengths.

    \textbf{Broader Impact.}
    By enabling anomaly detection on noisy, sparse, and irregular multivariate 
    time series, \gls{lsd} broadens the scope of deployable monitoring systems 
    for industrial applications, where early fault detection directly improves 
    safety, reliability, and sustainability. In line with the EU AI Act, we 
    advocate deploying \gls{lsd} as a decision-support tool, preserving operator 
    expertise and maintaining human oversight. Adversarial robustness remains an 
    open challenge and a promising direction for future work.

    \begin{ack}
        The financial support by the Austrian Federal Ministry of Economy, Energy
        and Tourism, the National Foundation for Research, Technology and
        Development and the Christian Doppler Research Association is gratefully
        acknowledged.
    \end{ack}

    \vskip3ex
    \begin{center}
    \begin{tikzpicture}
        \node[draw=none, rounded corners, fill=black!07!white, inner sep=5pt] at (0,0) {
        \textbf{Source code} is available at \url{https://github.com/plus-rkwitt/LatentSDEonHS}.
        };
    \end{tikzpicture}
    \end{center}
    \vspace{-1ex}

    \section*{References}
    \phantomsection
    \label{sec:references}
    \bibliographystyle{plainnat}
    \renewcommand{\bibsection}{}
    \bibliography{references}

    \appendix
    \newpage
    \begin{bibunit}
        \input{appendices/appendix}

        \newpage
        \section*{Appendix References}
        \putbib[references]
    \end{bibunit}

    \makeatletter
    \if@preprint
    \else
        \newpage
        \input{appendices/checklist}
    \fi
    \makeatother

\end{document}

%% file: appendices/appendix.tex
\appendix
\begin{center}
    \hrule height 4pt \vskip 0.25in
    {\LARGE\bfseries Supplementary Material \par}
    \vskip 0.29in \hrule height 1pt \vskip 0.09in
    \vskip 1em
\end{center}
This supplementary material provides additional details supporting the main 
paper. Appendix~\ref{app:appendix-benchmarks} gives detailed descriptions and 
statistics of all benchmark datasets. Appendix~\ref{app:appendix-detailed-experimental-settings} 
documents the full experimental setup, including data preparation, 
hyperparameter configurations, baseline implementations, and computing 
infrastructure. Runtime measurements for all methods are reported in 
Appendix~\ref{app:appendix-runtimes}.

\section{Benchmark Datasets}\label{app:appendix-benchmarks}

\Cref{tab:app-stat-benchmarks} reports the statistics of all benchmark datasets, 
including the number of individual traces, the total number of training and test 
samples, and the anomaly ratio on the test split.

\begin{table}[ht]
    \centering
    \caption{Statistics of the used benchmarks.}
    \label{tab:app-stat-benchmarks}
    \resizebox{\textwidth}{!}{
    \begin{tabular}{lccrrr}
        \toprule
        \textbf{Benchmark} & \textbf{\# Traces} & \textbf{\# Features} & \textbf{Total Train points} & \textbf{Total Test points} & \textbf{Total Anomaly Ratio} \\
        \midrule
        SWaT & 1 & 51 & 496800 & 449919 & 11.97\% \\
        WaDi & 1 & 130 & 784570 & 172803 & 5.77\% \\
        PSM & 1 & 25 & 132481 & 87841 & 27.76\% \\
        SMAP & 55 & 25 & 140825 & 444035 & 12.83\% \\
        MSL & 27 & 55 & 58317 & 73729 & 10.48\% \\
        SMD & 28 & 38 & 708405 & 708420 & 4.16\% \\
        QAPPD & 16 & 15 & 2912016 & 2912006 & 4.90\% \\
        \bottomrule
    \end{tabular}
    }
\end{table}

The benchmark datasets are publicly available at the following sources:
\begin{itemize}
    \item \gls{swat}~\citep{goh2017} and \gls{wadi}~\citep{chuadhry2017} are 
    published by iTrust, Centre for Research in Cyber Security, Singapore 
    University of Technology and Design; access requires 
    registration.\footnote{\url{https://itrust.sutd.edu.sg/itrust-labs_datasets/dataset_info/}}

    \item \gls{psm}~\citep{abdulaal2021} contains multivariate telemetry from 
    internal server infrastructure at eBay Inc., published under the 
    CC~BY~4.0 license.\footnote{\url{https://github.com/eBay/RANSynCoders}}

    \item \gls{msl} and \gls{smap}~\citep{hundman2018} contain telemetry from 
    NASA spacecraft missions and are made available by the authors under no 
    explicit license.\footnote{%
    \url{https://pds-atmospheres.nmsu.edu/data_and_services/atmospheres_data/Mars/Mars.html},
    \url{https://nsidc.org/data/smap}}

    \item \gls{smd}~\citep{su2019} contains server machine telemetry from a 
    large internet company and is available under the MIT 
    License.\footnote{\url{https://github.com/NetManAIOps/OmniAnomaly}}

    \item \gls{qad}~\citep{nosrati2026} is a recently introduced benchmark 
    of quasi-cyclic trajectories in state space, derived from a real 
    cyber-physical system, released under the CC~BY-SA~4.0 
    license.\footnote{\url{https://github.com/JRC-ISIA/industrial-federated-learning}} 
\end{itemize}

\section{Experimental Setup}
\label{app:appendix-detailed-experimental-settings}

\paragraph{Data Preparation and Splits.}
We follow a protocol similar to that proposed by~\cite{cai2026}. Each benchmark
dataset consists of a training set and a test set. The training set contains
only benign samples and is used for model fitting, whereas the test set contains
both benign and anomalous samples and is used to evaluate anomaly detection
performance. For methods that require a validation set, we randomly reserve
$10\%$ of the training data for validation and use the remaining $90\%$ for
training. For methods that do not require a validation set, \eg \gls{pca}, we
use the full training set for model fitting. 

Following~\cite{cai2026}, all
methods that require windowed inputs use a window size of $100$ and a stride of
$100$, resulting in non-overlapping windows across all benchmarks.
For the validation set, $10\%$ of the full windows are sampled.

\paragraph{Hyperparameter Settings.} We do not manually tune hyperparameters for the baseline methods. Instead, we use the default settings provided by the respective implementations, which already yield competitive performance across all benchmarks.

\begin{table}[ht]
    \centering
    \caption{The set hyperparameter for each of the benchmarks.}
    \label{tab:hyperparameter}
    \begin{tabular}{l|ccccccc}
        \toprule
        \textbf{Parameter} & \textbf{MSL} & \textbf{PSM} & \textbf{QAPPD} & \textbf{SMAP} & \textbf{SMD} & \textbf{SWaT} & \textbf{WaDi} \\
        \midrule
        Max. Epochs & 690 & 2000 & 690 & 690 & 690 & 2000 & 2000 \\
        Batch Size & 512 & 512 & 1024 & 512 & 512 & 256 & 128 \\
        Learning Rate & 0.001 & 0.1 & 0.05 & 0.001 & 0.001 & 0.05 & 0.05 \\
        Latent Dim. (z) & 16 & 8 & 4 & 16 & 12 & 16 & 14 \\
        Hidden Dim. (h) & 26 & 14 & 12 & 26 & 18 & 24 & 56 \\
        Degree (interpolation) & 12 & 6 & 6 & 12 & 8 & 12 & 16 \\
        Decoder Hidden Dim. & 24 & 12 & 11 & 24 & 16 & 22 & 64 \\
        KL$_0$ Weight & 0.0001 & 0.0001 & 0.001 & 0.0001 & 0.0001 & 0.0001 & 0.0001 \\
        KL$_p$ Weight & 0.001 & 0.001 & 0.01 & 0.001 & 0.001 & 0.001 & 0.001 \\
        Likelihood Weight & 1 & 1 & 100.0 & 1 & 1 & 1 & 100.0 \\
        $\sigma$ & 0.05 & 0.2 & 0.2 & 0.05 & 0.075 & 0.2 & 0.2 \\
        Subsample Ratio & 0.4   & 0.5 & 0.4 & 0.4 & 0.4 & 0.5 & 0.5 \\
        Normalize Score & $\checkmark$ & $\checkmark$ & $\checkmark$ & $\checkmark$ & $\checkmark$ & $\times$ & $\times$ \\
        \bottomrule
        \end{tabular}
\end{table}

\paragraph{Baseline Implementation.} 
Classical baseline methods (\gls{knn}, \gls{lof}, \gls{iforest}, \gls{ocsvm}, 
\gls{copod}, and \gls{pca}) are implemented using the PyOD~\citep{zhao2019} 
library\footnote{\url{https://pyod.readthedocs.io/en/latest/index.html}}. 
Deep learning-based baselines (DeepSVDD, USAD, TcnED, TranAD, 
AnomalyTransformer, DeepIF, TimesNet, and COUTA) are implemented using the 
DeepOD~\citep{xu2023} library\footnote{\url{https://deepod.readthedocs.io/en/latest/}}.

\paragraph{Method Implementation.}
\gls{lsd} builds on the implementation of \cite{zeng2023}, publicly available
online\footnote{\url{https://github.com/plus-rkwitt/LatentSDEonHS}}.
The architecture for the \gls{lsd} in particular consists of the following parts:

\begin{itemize}
    \item The \textbf{Recognition Network} maps the time-indexed 
    sequence of observations to the parameters of the approximate posterior
    of the \gls{sde}, implemented as a \gls{mtan}~\citep{shukla2021}.

    \item The \textbf{Encoder} integrates the latent path and is implemented 
    in two variants:
    \texttt{GLnPathDistributionEncoder} for \Rn and 
    \texttt{SOnPathDistributionEncoder} for \Sn.
    Both variants are parametrized by the latent dimension of $\mathbf{z}$ and the 
    degree of the Chebyshev polynomials $n$ used to approximate the drift function.

    \item The \textbf{Decoder} is a two-layer MLP with a ReLU activation on the 
    hidden layer. The hidden layer dimensionality is dataset dependent.
\end{itemize}

All parameters are optimized using Adam with cosine annealing learning rate 
scheduling. The training objective is the \gls{elbo}, with individually
weighted terms for the reconstruction likelihood, the 
\gls{kl} divergence of the initial latent state KL$_0$, and the 
divergence of the latent path samples KL$_p$.
The likelihood is evaluated under a normal distribution with a fixed 
standard deviation $\sigma$, treated as a hyperparameter.

\paragraph{Computing Infrastructure.} All \gls{lsd} experiments, except those 
on \gls{qad}, were conducted on a server equipped with an Intel Xeon Silver 
4114 CPU (2.20\,GHz) and NVIDIA GeForce RTX\,2080\,Ti GPUs (12\,GB VRAM). 
\gls{lsd} experiments on \gls{qad} were run on a server with an Intel Xeon 
E5-2687W v4 CPU (3.00\,GHz) and NVIDIA Titan\,X (Pascal) GPUs (12\,GB VRAM). 
All baseline experiments were conducted on a server with an Intel Core 
i9-10940X CPU (3.30\,GHz) and NVIDIA GeForce RTX\,3090 GPUs (24\,GB VRAM).

\section{Experimental Runtimes}\label{app:appendix-runtimes}

    \Cref{tab:app-runtimes} reports the runtime of a single training run for all baseline methods 
    and our proposed \gls{lsd} approach, in both \Rn and \Sn variants, across all seven benchmarks. 
    The reported values reflect training time only on a random seed, excluding data loading 
    overhead. For benchmarks comprising multiple traces, runtimes are aggregated over all 
    traces.
    
    The experimental runtimes show that \gls{lsd} incurs substantially higher training
    times than all baseline methods. This is expected given the complexity of fitting 
    a latent \gls{sde} model and the iterative nature of numerical solvers required 
    for training. However, once trained, the inference time of \gls{lsd} is comparable
    to that of deep learning-based baselines, as anomaly scoring only requires a single
    forward pass through the decoder.

    Thus, the increased training cost is a one-time overhead. We consider this 
    trade-off reasonable given the observed performance gains, particularly in 
    settings with sparse and irregularly sampled data.

    \begin{table}
        \caption{Average training runtimes (in minutes) per run for \gls{lsd} and all 
            baseline methods. Runtimes are measured on a single GPU where applicable.}
        \label{tab:app-runtimes}
        \resizebox{\textwidth}{!}{
        \begin{tabular}{lrrrrrrr}
        \toprule
        \textbf{Method}             & \textbf{SWaT (1)} & \textbf{WaDi (1)} & \textbf{PSM (1)} & \textbf{SMAP (55)} & \textbf{MSL (27)} & \textbf{SMD (28)} & \textbf{QAPPD (16)} \\
        \midrule
        COPOD & 0.29 & 0.96 & 0.05 & 0.21 & 0.11 & 0.35 & 0.94 \\
        IForest & 0.10 & 0.29 & 0.03 & 0.29 & 0.15 & 0.32 & 1.00 \\
        KNN & 2.74 & 7.52 & 0.16 & 0.18 & 0.11 & 0.51 & 0.97 \\
        LOF & 2.82 & 7.38 & 0.16 & 0.21 & 0.11 & 0.50 & 1.17 \\
        OCSVM & 510.48 & 2190.78 & 21.83 & 0.75 & 0.33 & 28.44 & 5.65 \\
        PCA & 0.07 & 0.27 & 0.02 & 0.13 & 0.10 & 0.17 & 0.90 \\
        \midrule
        AnomalyTransformer & 3.83 & 4.60 & 0.96 & 3.82 & 0.96 & 7.76 & 4.01 \\
        COUTA & 27.67 & 42.28 & 7.34 & 8.16 & 3.42 & 35.91 & 15.20 \\
        DeepIF & 30.11 & 67.32 & 5.67 & 13.81 & 5.59 & 42.67 & 14.93 \\
        DeepSVDD & 1.73 & 2.72 & 0.41 & 1.19 & 0.55 & 2.65 & 1.98 \\
        TcnED & 3.33 & 3.05 & 0.60 & 2.46 & 0.61 & 8.67 & 3.36 \\
        TimesNet & 10.16 & 11.81 & 2.50 & 6.66 & 1.82 & 18.54 & 8.41 \\
        TranAD & 1.21 & 3.07 & 0.19 & 0.74 & 0.32 & 1.72 & 1.41 \\
        USAD & 1.08 & 5.58 & 0.15 & 1.78 & 1.61 & 2.07 & 1.65 \\
        \midrule
        LSD on \Rn & 156 & 116 & 184 & 804 & 411 & 487 & 835 \\
        LSD on \Sn & 196 & 196 & 173 & 1148 & 371 & 1066 & 840 \\
        \bottomrule
        \end{tabular}
        }
    \end{table}

%% file: appendices/checklist.tex
\section*{NeurIPS Paper Checklist}

\begin{enumerate}

\item {\bf Claims}
    \item[] Question: Do the main claims made in the abstract and introduction accurately reflect the paper's contributions and scope?
    \item[] Answer: \answerYes{}.
    \item[] Justification: All claims stated in the abstract and introduction are directly supported by the theoretical and experimental results presented in the paper. The contributions, assumptions, and scope are explicitly described, and the level of generalization is aligned with the empirical evidence and discussed limitations.

\item {\bf Limitations}
    \item[] Question: Does the paper discuss the limitations of the work performed by the authors?
    \item[] Answer: \answerYes{}.
    \item[] Justification: The paper includes a dedicated discussion of limitations in \Cref{sec:discussion-limitations}, where we explicitly outline key assumptions, potential failure modes, and factors influencing performance. We also discuss the scope of empirical validation, including dataset and experimental constraints, and comment on the expected generalization and computational considerations of the proposed approach.

\item {\bf Theory assumptions and proofs}
    \item[] Question: For each theoretical result, does the paper provide the full set of assumptions and a complete (and correct) proof?
    \item[] Answer: \answerNA{}.
    \item[] Justification: The paper does not include theoretical results; all contributions are empirical, and therefore no assumptions or proofs are required.

\item {\bf Experimental result reproducibility}
    \item[] Question: Does the paper fully disclose all the information needed to reproduce the main experimental results of the paper to the extent that it affects the main claims and/or conclusions of the paper (regardless of whether the code and data are provided or not)?
    \item[] Answer: \answerYes{}
    \item[] Justification: All information required to reproduce the main experimental results is provided in the paper and supplementary material, including detailed descriptions of the methodology, experimental setup, and evaluation protocols. The approach builds on established methods, which are properly referenced, and all benchmark datasets used are publicly available. In addition, the full implementation will be released upon acceptance to further facilitate reproducibility.

    \item[] Guidelines:
    \begin{itemize}
        \item If the paper includes experiments, a \answerNo{} answer to this question will not be perceived well by the reviewers: Making the paper reproducible is important, regardless of whether the code and data are provided or not.
        \item If the contribution is a dataset and\slash or model, the authors should describe the steps taken to make their results reproducible or verifiable. 
        \item Depending on the contribution, reproducibility can be accomplished in various ways. For example, if the contribution is a novel architecture, describing the architecture fully might suffice, or if the contribution is a specific model and empirical evaluation, it may be necessary to either make it possible for others to replicate the model with the same dataset, or provide access to the model. In general. releasing code and data is often one good way to accomplish this, but reproducibility can also be provided via detailed instructions for how to replicate the results, access to a hosted model (e.g., in the case of a large language model), releasing of a model checkpoint, or other means that are appropriate to the research performed.
        \item While NeurIPS does not require releasing code, the conference does require all submissions to provide some reasonable avenue for reproducibility, which may depend on the nature of the contribution. For example
        \begin{enumerate}
            \item If the contribution is primarily a new algorithm, the paper should make it clear how to reproduce that algorithm.
            \item If the contribution is primarily a new model architecture, the paper should describe the architecture clearly and fully.
            \item If the contribution is a new model (e.g., a large language model), then there should either be a way to access this model for reproducing the results or a way to reproduce the model (e.g., with an open-source dataset or instructions for how to construct the dataset).
            \item We recognize that reproducibility may be tricky in some cases, in which case authors are welcome to describe the particular way they provide for reproducibility. In the case of closed-source models, it may be that access to the model is limited in some way (e.g., to registered users), but it should be possible for other researchers to have some path to reproducing or verifying the results.
        \end{enumerate}
    \end{itemize}

\item {\bf Open access to data and code}
    \item[] Question: Does the paper provide open access to the data and code, with sufficient instructions to faithfully reproduce the main experimental results, as described in supplemental material?
    \item[] Answer: \answerYes{}.
    \item[] Justification: Six out of seven benchmark datasets used in the study are publicly available; their sources, access procedures, and licenses are documented in Appendix~\ref{app:appendix-benchmarks}, enabling transparency and reproducibility. The remaining dataset is not yet publicly released, but will be during the review period; for the purpose of the review process, we provide a representative subset in the supplemental material.

\item {\bf Experimental setting/details}
    \item[] Question: Does the paper specify all the training and test details (e.g., data splits, hyperparameters, how they were chosen, type of optimizer) necessary to understand the results?
    \item[] Answer: \answerYes{}.
    \item[] Justification: All relevant training and evaluation details are specified in \Cref{sec:empirical-evaluation} and Appendix~\ref{app:appendix-detailed-experimental-settings}, including dataset splits, preprocessing, model configurations, hyperparameters and their selection, optimization procedures, and evaluation protocols, ensuring that the results are fully transparent and interpretable.

\item {\bf Experiment statistical significance}
    \item[] Question: Does the paper report error bars suitably and correctly defined or other appropriate information about the statistical significance of the experiments?
    \item[] Answer: \answerYes{}.
    \item[] Justification: All experiments are conducted over multiple independent 
    runs, with results reported as mean$\pm$standard deviation, reflecting 
    variability arising from random initialization and stochastic training. 
    To enable a metric- and scale-free comparison across all evaluated methods 
    and benchmarks, we additionally report method rankings aggregated across 
    all benchmarks and metrics.

\item {\bf Experiments compute resources}
    \item[] Question: For each experiment, does the paper provide sufficient information on the computer resources (type of compute workers, memory, time of execution) needed to reproduce the experiments?
    \item[] Answer: \answerYes{}.
    \item[] Justification: All relevant computational details are provided in Appendix~\ref{app:appendix-detailed-experimental-settings}, including the type of hardware used (CPU/GPU), memory specifications, and runtime characteristics of the experiments. We also report the approximate compute requirements per experimental run and the overall compute budget required for the reported results, ensuring reproducibility and transparency.
    
\item {\bf Code of ethics}
    \item[] Question: Does the research conducted in the paper conform, in every respect, with the NeurIPS Code of Ethics \url{https://neurips.cc/public/EthicsGuidelines}?
    \item[] Answer: \answerYes{}.
    \item[] Justification: We have carefully reviewed the NeurIPS Code of Ethics and confirm that all aspects of the research, including data usage, experimental procedures, and reporting, fully comply with these guidelines.

\item {\bf Broader impacts}
    \item[] Question: Does the paper discuss both potential positive societal impacts and negative societal impacts of the work performed?
    \item[] Answer: \answerNA{}.
    \item[] Justification: The proposed work focuses on anomaly detection methods evaluated on publicly available benchmark datasets, without introducing new data sources or deployment-specific systems. While the methodological framing addresses practical challenges encountered in real-world settings, it does not create a direct pathway to applications with foreseeable societal risks. Given the general-purpose nature of anomaly detection and the absence of sensitive data or high-stakes deployment scenarios, we do not identify meaningful positive or negative societal impacts arising specifically from this work.
    
\item {\bf Safeguards}
    \item[] Question: Does the paper describe safeguards that have been put in place for responsible release of data or models that have a high risk for misuse (e.g., pre-trained language models, image generators, or scraped datasets)?
    \item[] Answer: \answerNA{}.
    \item[] Justification: No pre-trained models or datasets are released as part of this work, and therefore no specific safeguards for model or data misuse are required.

\item {\bf Licenses for existing assets}
    \item[] Question: Are the creators or original owners of assets (e.g., code, data, models), used in the paper, properly credited and are the license and terms of use explicitly mentioned and properly respected?
    \item[] Answer:\answerYes{}.
    \item[] Justification: All existing assets used in this work, including benchmark datasets and baseline methods, are properly credited through citations to the original sources. For each benchmark dataset, we additionally provide the corresponding license information and links to the official data sources, ensuring that all terms of use are explicitly stated and respected.

\item {\bf New assets}
    \item[] Question: Are new assets introduced in the paper well documented and is the documentation provided alongside the assets?
    \item[] Answer: \answerYes{}.
    \item[] Justification: A link to the code repository is provided in the paper, and the implementation is included in anonymized form within the supplemental material for peer review. The provided code includes sufficient documentation to reproduce the proposed methods and experimental results.

\item {\bf Crowdsourcing and research with human subjects}
    \item[] Question: For crowdsourcing experiments and research with human subjects, does the paper include the full text of instructions given to participants and screenshots, if applicable, as well as details about compensation (if any)? 
    \item[] Answer: \answerNA{}.
    \item[] Justification: This research does not involve crowdsourcing, human subjects, or any form of user studies; therefore, no participant instructions, interfaces, or compensation details are applicable.

\item {\bf Institutional review board (IRB) approvals or equivalent for research with human subjects}
    \item[] Question: Does the paper describe potential risks incurred by study participants, whether such risks were disclosed to the subjects, and whether Institutional Review Board (IRB) approvals (or an equivalent approval/review based on the requirements of your country or institution) were obtained?
    \item[] Answer: \answerNA{}.
    \item[] Justification: This research does not involve human subjects, crowdsourcing, or user studies; therefore, no IRB approval (or equivalent ethical review) was required.

\item {\bf Declaration of LLM usage}
    \item[] Question: Does the paper describe the usage of LLMs if it is an important, original, or non-standard component of the core methods in this research? Note that if the LLM is used only for writing, editing, or formatting purposes and does \emph{not} impact the core methodology, scientific rigor, or originality of the research, declaration is not required.
    \item[] Answer: \answerNA{}.
    \item[] Justification: The core methodology of this work does not involve the use of LLMs as a component of the proposed approach. Any use of LLMs, if present, is limited to auxiliary purposes (e.g., writing or formatting) and does not affect the scientific methods, experiments, or conclusions of the paper.

\end{enumerate}